\pdfoutput=1

\documentclass[letterpaper, 10 pt, conference]{ieeeconf}  % Comment this line out if you need a4paper

\IEEEoverridecommandlockouts                              % This command is only needed if 
\usepackage{graphics} % for pdf, bitmapped graphics files
\usepackage{amsmath,amssymb,amsfonts} % assumes amsmath package installed
\usepackage{bm}
\usepackage{subcaption}	 
\usepackage{graphicx}
\usepackage{multirow}
\usepackage{array}
\usepackage{url}
\usepackage{booktabs}
\usepackage{hyperref}
\hypersetup{colorlinks,urlcolor=red}
\usepackage{xcolor}
\usepackage{bbold}

\usepackage[backend=bibtex,natbib=true,style=ieee,sorting=none,giveninits=true,maxbibnames=99,url=false,doi=false]{biblatex}
\title{\LARGE \bf
Monocular Event-Inertial Odometry with Adaptive decay-based Time Surface and Polarity-aware Tracking
}

\author{Kai Tang, Xiaolei Lang, Yukai Ma, Yuehao Huang, Laijian Li, Yong Liu{$^{\ast}$}, Jiajun Lv{$^{\ast}$} % <-this % stops a space
\thanks{The authors are with the Institute of Cyber-Systems and Control, Zhejiang University, Hangzhou, China. This work was supported by NSFC 62088101 Autonomous Intelligent Unmanned Systems. }
\thanks{
{$^\ast$} Yong Liu and Jiajun Lv are the corresponding authors, email: yongliu@iipc.zju.edu.cn, lvjiajun314@zju.edu.cn}
}

\begin{document}

\maketitle
\renewcommand{\thefootnote}{\fnsymbol{footnote}}
 % \footnotetext{* These authors contributed equally to this work.}
\thispagestyle{empty}
\pagestyle{empty}

%%%%%%%%%%%%%%%%%%%%%%%%%%%%%%%%%%%%%%%%%%%%%%%%%%%%%%%%%%%%%%%%%%%%%%%%%%%%%%%%
\begin{abstract}
% 事件相机因其相较于普通相机具有无动态模糊和动态范围大的优势，逐渐成为研究领域的一项热点。为了从事件中获取信息，目前被广泛使用的是一种基于指数衰减核的时间面表示。但由于其中的时间常数通常需事先设定，且不能随事件流的动态特性发生变化，从而使生成的时间面很难清晰地反应环境纹理。本文中，我们提出了一种基于事件的视觉惯性里程计，其中采用了基于自适应衰减核的时间面表示，里程计后端采用了多状态约束卡尔曼滤波方法。时间面若采用极性进行加权时，在自身运动方向突然发生变化时容易因为事件的极性变化，出现特征跟踪丢失的问题，为此我们优化了KLT的跟踪过程以提升了特征跟踪的鲁棒性。在本文的实验部分，我们和一些基于帧和基于事件的视觉惯性里程计方法在精度上进行了对比，并在消融实验中对比了两种时间面和及其不同参数选择下的里程计精度。实验结果表明，我们的方法在上述测试中均能够取得十分有竞争力的结果。
Event cameras have garnered considerable attention due to their advantages over traditional cameras in low power consumption, high dynamic range, and no motion blur. 
This paper proposes a monocular event-inertial odometry incorporating an adaptive decay kernel-based time surface with polarity-aware tracking. 
We utilize an adaptive decay-based Time Surface to extract texture information from asynchronous events, which adapts to the dynamic characteristics of the event stream and enhances the representation of environmental textures. 
However, polarity-weighted time surfaces suffer from event polarity shifts during changes in motion direction. To mitigate its adverse effects on feature tracking, we optimize the feature tracking by incorporating an additional polarity-inverted time surface to enhance the robustness. 
Comparative analysis with visual-inertial and event-inertial odometry methods shows that our approach outperforms state-of-the-art techniques, with competitive results across various datasets.

\end{abstract}

% Main sections
\section{Introduction}

Accurate environmental perception is essential in robotics. Traditional vision sensors, like conventional cameras, often suffer from motion blur during rapid movement and can lose image details due to their limited dynamic range, thereby undermining perception accuracy and robustness. Event cameras, equipped with Dynamic Vision Sensors (DVS), offer a promising solution to these challenges. They generate events asynchronously whenever a pixel's brightness change surpasses a preset threshold. Consequently, event cameras boast a wide dynamic range, high temporal resolution, low energy consumption, and immunity to motion blur.

It is commonly recognized that high-textured regions trigger events more frequently than low-textured ones, making it possible to extract texture details from the event stream. However, processing these asynchronous events is a challenging task. For this reason, various event representation methods have been proposed, and Gallego et al.~\cite{gallego2020event} classify these event representations into Individual Events, Event Packet, Event Frame/Image, Time Surface, etc. Time Surface~\cite{delbruck2008frame}, a 2D map representation in which each pixel stores the timestamp of the corresponding event, is widely utilized in event-based SLAM or odometry methods~\cite{zhou2021stereo,zuo2022devo}. 

\begin{figure}[t]
    \centering
    \includegraphics[width=\linewidth]{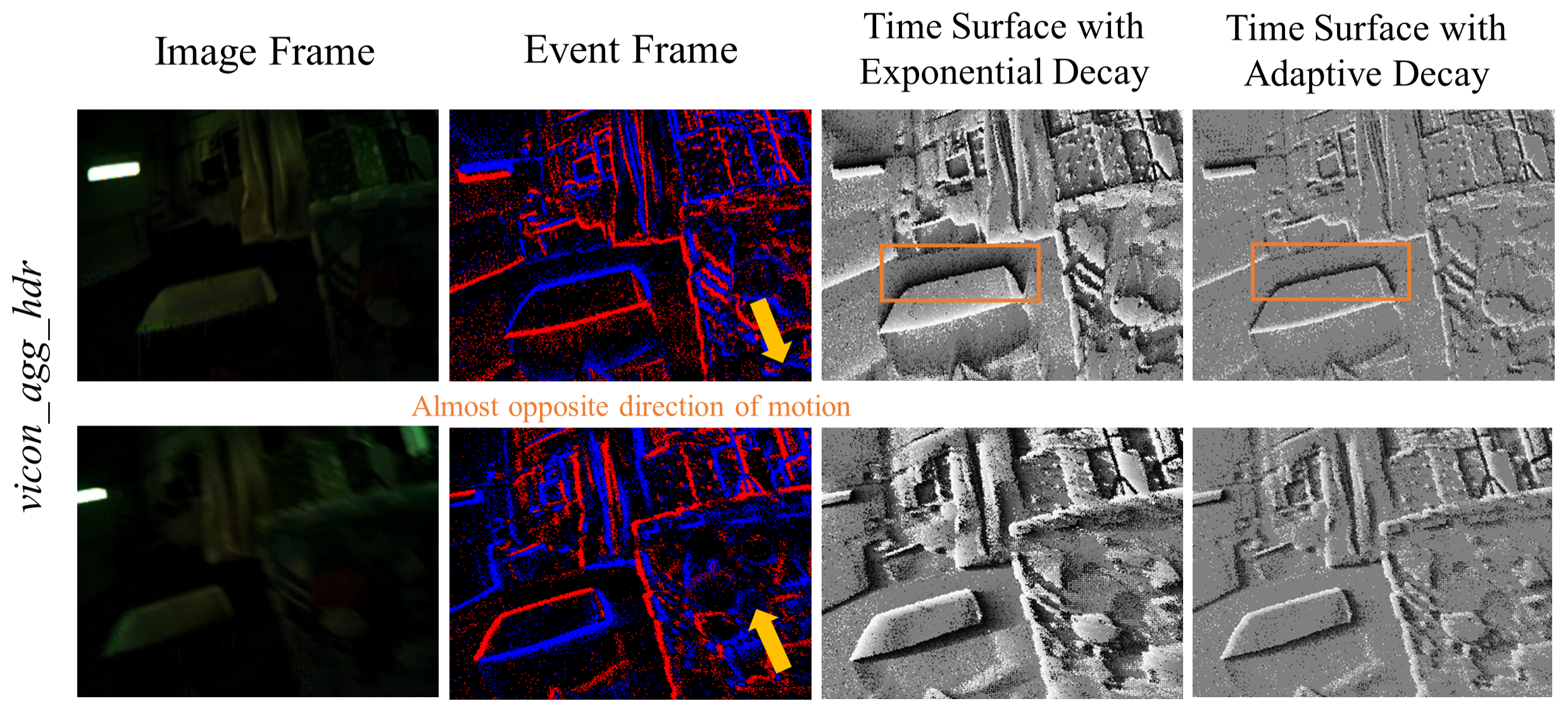}
    \caption{Event cameras have potential advantages over traditional cameras in high dynamic range and high-speed motion scenes. Time surfaces with an exponential decay kernel (the third column) are unable to adjust their internal parameter (time constant $\eta$) in response to the event stream's dynamic characteristics, resulting in bold edges and redundant events. Conversely, adaptive decay-based time surfaces (the fourth column) offer clear details and less noise.}
    \label{fig:sec1:event_frame}
    \vspace{-1em}
\end{figure}

To highlight recent events over past events, time surfaces often employ an exponential decay kernel~\cite{lagorce2016hots}, but this method presents some inherent limitations. Initially, the exponential decay kernel requires a preset parameter—time constant $\eta$, which requires manual adjustment for different sequences. It lacks adaptability to the dynamics of the event stream, and this approach is prone to bold edges and noticeable trailing when the event frequency is high. Moreover, it does not filter out events that contribute less to the texture, resulting in superfluous events within time surfaces.
Furthermore, the motion direction of the event camera influences the polarities. When sudden changes occur in the direction of motion, the most recent events may exhibit opposite polarities. Consequently, this can lead to fluctuations in the grayscale values of the pixels associated with these events, potentially disrupting feature tracking.

To tackle the aforementioned challenges, inspired by~\cite{nunes2023adaptive}, we propose a monocular event-based visual inertial odometry with the adaptive decay-based time surface representation and polarity-aware tracking. 
The main contributions of the paper are concluded as follows:
\begin{itemize}
    \item We propose a real-time monocular event-inertial odometry based on adaptive decay-based time surface and polarity-aware tracking within the MSCKF framework for accurate pose estimation.
    \item We present an adaptive decay-based time surface to accommodate the dynamic characteristics of the event stream. And we propose a polarity-aware tracking method that improves the stability of feature tracking by utilizing an additional polarity-inverted time surface.
    \item We evaluate the proposed method using different datasets, comparing its accuracy with visual-inertial odometry and event-inertial odometry methods, and assessing the efficiency of the adaptive decay and the proposed tracking approach. Experimental results show that our method produces competitive results.
\end{itemize}

The remainder of this paper is organized as follows: Section \ref{sec:related work} provides a brief summary of recent works. Our system is described in Section \ref{sec:method}. Section \ref{section:experiments} presents detailed experimental settings and results in multiple datasets. Finally, Section \ref{sec:conclusion} offers a succinct overview of our system and outlines future directions.

\section{Related Works}
\label{sec:related work}

\subsection{Event-based Visual Odometry}
Kueng et al.~\cite{kueng2016lowlatency} developed a pioneering method for event-based visual odometry that relies on feature extraction from grayscale maps and asynchronous event-based tracking, facilitating 6DoF pose estimation and mapping. However, this method is confined to feature points detected in image frames. Another stride in event-based visual odometry by Kim et al.~\cite{kim2016real} achieved real-time event-based SLAM by integrating triplet probabilistic filters for pose estimation, scene mapping, and intensity estimation, although this approach demands GPU acceleration due to its computational intensity.
EVO~\cite{rebecq2016evo} advanced the field by presenting an image-to-model alignment-based tracking approach to capture rapid camera movements, recovering semi-dense maps in parallel processing, although a lengthy start-up phase and poor accuracy hinder it. EDS~\cite{hidalgo-carrio2022eventaided} improved upon this with an event generation model to track camera motion, enhancing the accuracy of visual odometry but at the expense of increased computational demand and slower optimization speeds.
Lastly, an innovative 6DoF motion compensation mechanism by Huang et al.~\cite{huang2023mcveo} allowed deblurred event frame generation synchronized to RGB images using an event generation model, addressing the modality disparities between images and event data.

Geometric approaches inadequately harness event data, whereas deep learning-based event odometers introduce an innovative processing paradigm. RAMP-VO~\cite{pellerito2024endtoendlearnedvisualodometry} is the inaugural end-to-end framework for event- and image-based visual odometry, merging asynchronous event streams with image data through a parallel encoding scheme. Alternatively, DEVO~\cite{klenk2023deep} constitutes the premier monocular event-only odometry system, grounded in deep learning, which trains on event voxel grids and inverse depth maps under the guidance of ground truth poses, achieving markedly enhanced accuracy over conventional techniques. Although deep learning odometry signifies a stride in precision, cost-effective accuracy enhancements on computation-constrained platforms, such as drones, may still benefit from integrating event data with IMU measurements.

\subsection{Event-based Visual Inertial Odometry}

Zhu et al.~\cite{zhu2017eventbased} introduced the pioneering event-based odometry by integrating an event-driven tracker with IMU data. However, its real-time application is hindered by computationally demanding feature tracking.
Vidal et al.~\cite{vidal2018ultimate} advanced the field by establishing a tightly integrated framework combining events, frames, and inertial readings for enhanced state estimation.
Complementarily, some research integrates events and IMU within a continuous-time schema. Mueggler et al.~\cite{mueggler2018continuoustime} devised a continuous-time approach for merging high-frequency event and IMU data for visual inertial odometry. 
Embracing a sophisticated feature tracker~\cite{gehrig2020eklt}, EKLT-VIO~\cite{mahlknecht2022exploring} demonstrated accurate performance in Mars-like and high-dynamic-range sequences and showed good potential in vision-based exploration on Mars.
Dai et al.~\cite{dai2022tightlycoupled} presented a comprehensive model marrying continuous-time inertial data with events, innovating an exponential decay correlation for events.
Guan et al.~\cite{guan2022monocular} developed a uniformly distributed event corner detection algorithm for raw events, and designed two event representations to perform feature tracking and loop closure matching for a keyframe-based visual inertial system. 
Based on this, Guan et al.~\cite{guan2023plevio} utilized motion compensation for the event stream based on IMU measurements and introduced line and point feature constraints, improving odometric precision.

Although time surfaces based on exponential decay can be additionally motion compensated to enhance performance, as employed by \cite{huang2023mcveo, guan2023plevio}, they still rely on accurate sensor measurements (e.g. IMU) or accurate odometer estimates. To make the exponential decay kernel adaptable to the dynamics of the event stream, we adopted the event activity model proposed by Nunes et al.~\cite{nunes2023adaptive}, and introduced an adaptive decay kernel for different event cameras and practical operating conditions. Leveraging such advancements, we propose a monocular event-inertial odometry with the adaptive decay kernel and optimize feature tracking for the problems faced by the polarity-weighted time surface-based approach, resulting in a more robust and accurate odometry.

\section{Methodology}
\label{sec:method}

\begin{figure*}[t]
    \vspace{5pt}
    \centering
    \includegraphics[width=0.9\linewidth]{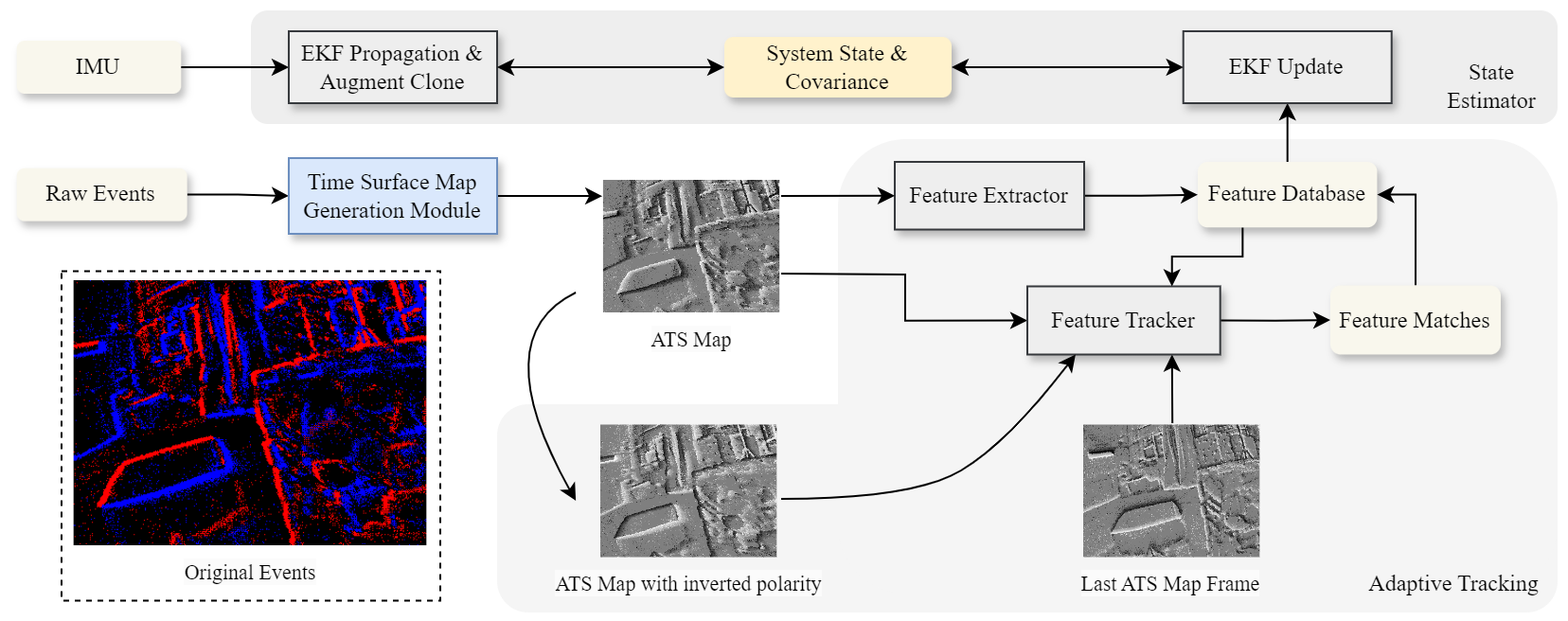}
    \caption{System overview. The system includes the Time Surface Map Generation Module, the Adaptive Tracking Module and the State Estimator. The Adaptive Tracking Module fuses the tracking results of Polarity-weighted and Polarity-inverted time surfaces to alleviate the tracking loss that occurs when the direction of motion or illumination changes.}
    \label{fig:sec3:system_overview}
    \vspace{-1em}
\end{figure*}

% \subsection{MSCKF-based Event{\color{blue}-inertial} Odometry}
\subsection{System Overview}

% pipline
The system overview is depicted in Fig.~\ref{fig:sec3:system_overview}. 
Raw events are transformed into adaptive decay-based time surfaces (Sec.~\ref{sec:ats}) using the proposed Time-Priority strategy. 
Data from a monocular event camera and an IMU are fused in  the Multi-State Constraint Kalman Filter (MSCKF)~\cite{mourikis2007multistate}, enabling precise and low-latency pose estimation.
Given the sparse event outputs from the event camera when the system is in static or slow motion, we utilize a dynamic initialization method~\cite{geneva2020openvins} to incorporate feature observations across time surfaces and IMU measurements for system initialization.
Once initialized, upon receiving a new event packet, the system leverages IMU measurements to propagate the mean and covariance of the state to the timestamp of the new event packet, and augments a cloned IMU pose to the state vector (Sec.~\ref{sec:state_vec}). Afterward, sparse feature observations from the proposed tracking method (Sec.~\ref{sec:feature_tracking}) are utilized to update the state (Sec.~\ref{sec:feature_udpate}). Landmarks and old cloned poses are marginalized out of the state vector for computational efficiency.

\subsection{State Vector of Event-Inertial Odometry} 
\label{sec:state_vec}

The state vector $\mathbf{x}_j$ at timestamp $t_j$ of the system is composed of the inertial state $\mathbf{x}_{I_j}$, inertial pose clones $\mathbf{x}_C$ and the inverse depths of landmarks $\mathbf{x}_f$, given by
\begin{equation}
    \label{eqn:state_vector}
    \begin{aligned}
        \mathbf{x}_j &= \left[\mathbf{x}_{I_j}^T \quad \mathbf{x}_C^T \quad \mathbf{x}_{f}^T \right]^T \,,\\
        \mathbf{x}_{I_j} &= \left[{}^{I_j}_G\overline{q}^T \quad {}^G\mathbf{p}_{I_j}^T \quad {}^G\mathbf{v}_{I_j}^T \quad \mathbf{b}_{a_j}^T \quad \mathbf{b}_{g_j}^T \right]^T \,,\\
        \mathbf{x}_{C} &= \left[{}^{I_{j}}_G\overline{q}^T \quad {}^G\mathbf{p}_{I_{j}}^T \quad \cdots \quad {}^{I_{j-m}}_G\overline{q}^T \quad {}^G\mathbf{p}_{I_{j-m}}^T \right]^T \\
        \mathbf{x}_{f} &= \left[\rho_1 \quad \cdots \quad \rho_l \quad \cdots  \quad \rho_s \right]^T \,,
    \end{aligned}
\end{equation}
where ${}^{I_j}_G\overline{q}$ denotes the unit quaternion, and the corresponding rotation matrix is denoted as $\mathbf{R}({}^{I_j}_G\overline{q})={}^{I_j}_G\mathbf{R}$. Furthermore, ${}^G\mathbf{p}_{I_j}$ and ${}^G\mathbf{v}_{I_j}$ denote the position and velocity of the IMU in the global frame at time $t_j$, while $\mathbf{b}_{a_j}$ and $\mathbf{b}_{g_j}$ are the biases of the accelerometer and the gyroscope, respectively. 

The sliding window maintains a set of $m+1$ cloned IMU poses at the timestamps of the event packet for feature triangulation and state update like an RGB-Camera-based VIO. Stable features tracked across the entire sliding window frames are considered as SLAM features and will be augmented to the state vector for extending the time span of active constraints. In this paper, we use the inverse depth parameterization and only include maximum $s$ landmarks in the state for limiting computational complexity. 
The extrinsic parameters ${}^C_I\mathbf{R}, {}^C\mathbf{p}_I$ between the IMU and the camera are precalibrated and assumed to be known. In our practical experiments, $m$ and $s$ are set to 10 and 50, respectively.

\subsection{Time Surface and Adaptive decay}
\label{sec:ats}

\smallskip
\noindent
\textbf{Event Definition:} 
Events primarily occur in regions with rich environmental textures, such as edges.
When the brightness of a pixel changes beyond a certain threshold, an event is triggered. An event at timestamp $t_k$ with pixel coordinate $[u_k, v_k]^T$ is defined as:
\begin{equation}
    e_k = \{u_k, v_k, t_k, p_k\}\,,
\end{equation}
where the polarity $p_k\in\{-1, 1\}$ indicates whether the brightness increase or decrease.

\smallskip
\noindent
\textbf{Time Surface:} 
As illustrated in Fig.~\ref{fig:sec1:event_frame}, event frames (in the second column) generated from spatio-temporal neighborhoods of raw event are the primitive method for representing event positions within an image. This representation method is highly sensitive to event counts and often struggles to accurately capture environmental texture due to the lack of consideration for event timestamps.
To address this issue, the Time Surface~\cite{delbruck2008frame}, which retains event timestamps, is proposed. To prioritize recent event information, the time surface is usually used together with an exponential decay kernel~\cite{lagorce2016hots}:
\begin{equation}
    \label{eqn:ts}
    \tau_1(e_p, t) \doteq \exp{(-\frac{t-t_{p}}{\eta})} \,,
\end{equation}
where $\eta$ represents a predefined time constant and $e_p$ is any previous event and $t_p$ is its timestamp. 
The time constant $\eta$ needs to be fine-tuned to mitigate interference from past events. However, this invariant time constant $\eta$ does not accommodate all camera motion and should be adjusted with different motion situations. A constant $\eta$ results in bold edges in the time surfaces when the motion is aggressive.

\smallskip
\noindent
\textbf{Adaptive Decay Kernel:} 
To reflect the dynamic characteristics of the event stream, \cite{nunes2023adaptive} proposes the \textit{Event Activity} $\alpha(t)$ as follows:
\begin{equation}
    \label{eqn:event_activity}
    \alpha(t) = \tau_2(e_{p}, t)\alpha(t_p) + n(t,t_{p}) \,,
\end{equation}
where $t_{p}$ is the timestamp of any previous event $e_p$ and $n(t,t_{p})$ is the count of events within the time interval $[t_p, t]$. Then we can obtain the adaptive decay kernel for the time surface, which is derived from the exponential decay. Given the time derivation of Eq.~\eqref{eqn:ts}: 
\begin{equation}
    \label{eqn:dts}
    \begin{aligned}
    \frac{\partial \tau_1(e_p, t)}{\partial t}
     &= - \frac{1}{\eta}\exp{(-\frac{t-t_p}{\eta})} = -\lambda(t) \tau_1(e_p,t)  \,,
    \end{aligned}
\end{equation}
where $\lambda$ represents the decay rate which we expect to be time-varying and influenced by the event activity. The decay rate $\lambda(t)$ is intended to be proportional to the event activity $\lambda(t) \propto \alpha(t)$, solving the Eq.~\eqref{eqn:dts} yields the \textit{adaptive decay}:
\begin{equation}
    \label{eqn:adaptive_decay_kernel}
    \tau_2(e_p, t) = \frac{1}{1+r\cdot\alpha(t_{p})(t-t_{p})} \,,
\end{equation}
where $\alpha(t_{p})$ is the event activity at previous timestamp $t_p$ and $r$ is the coefficient. Eq.~\eqref{eqn:adaptive_decay_kernel} introduces the coefficient $r$,  which differs from \cite{nunes2023adaptive} where the coefficient $r$ is set to a constant of 1. However, this setting ignores the influence of the camera resolution as well as the threshold for triggering events, resulting in too little decay for redundant events or too much decay for active events.
% Finding more details at~\cite{nunes2023adaptive} or our supplement material.

The number of event activities increases with event frequency. As inferred from Eq.~\eqref{eqn:adaptive_decay_kernel}, when the event frequency increases, events with the later timestamp have less decays. In other words, if the pixel value remains constant, a higher event frequency implies that the event timestamps are closer to $t$. Event information is retained for a longer period of time when the event frequency is high. Therefore, modeling the dynamics of the event stream by introducing the event activity enables adaptively adjusting the decay for the events and thus provides high-quality imaging.

\begin{figure}[t]
    \vspace{5pt}
    \centering
    \includegraphics[width=0.9\linewidth]{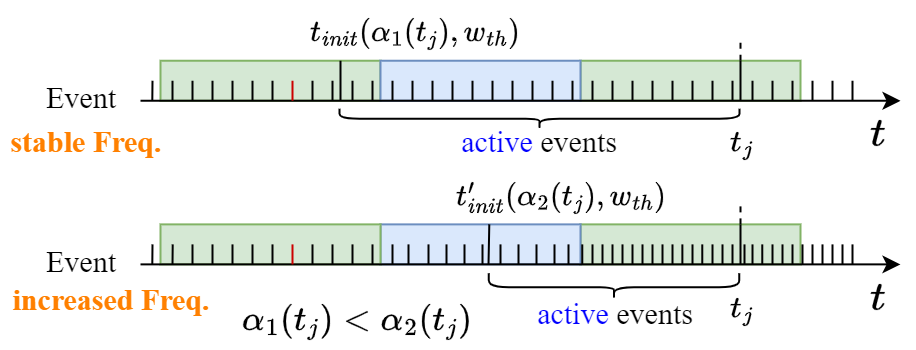}
    \caption{The impact of event activity on $t_{init}$. The increased value of event activity shortens the temporal scope of active events, thereby mitigating interference from past events.}
    \label{fig:sec3:influence_of_activity}
    \vspace{-1em}
\end{figure}

\begin{figure}[t]
    \vspace{5pt}
    \centering
    \includegraphics[width=0.85\linewidth]{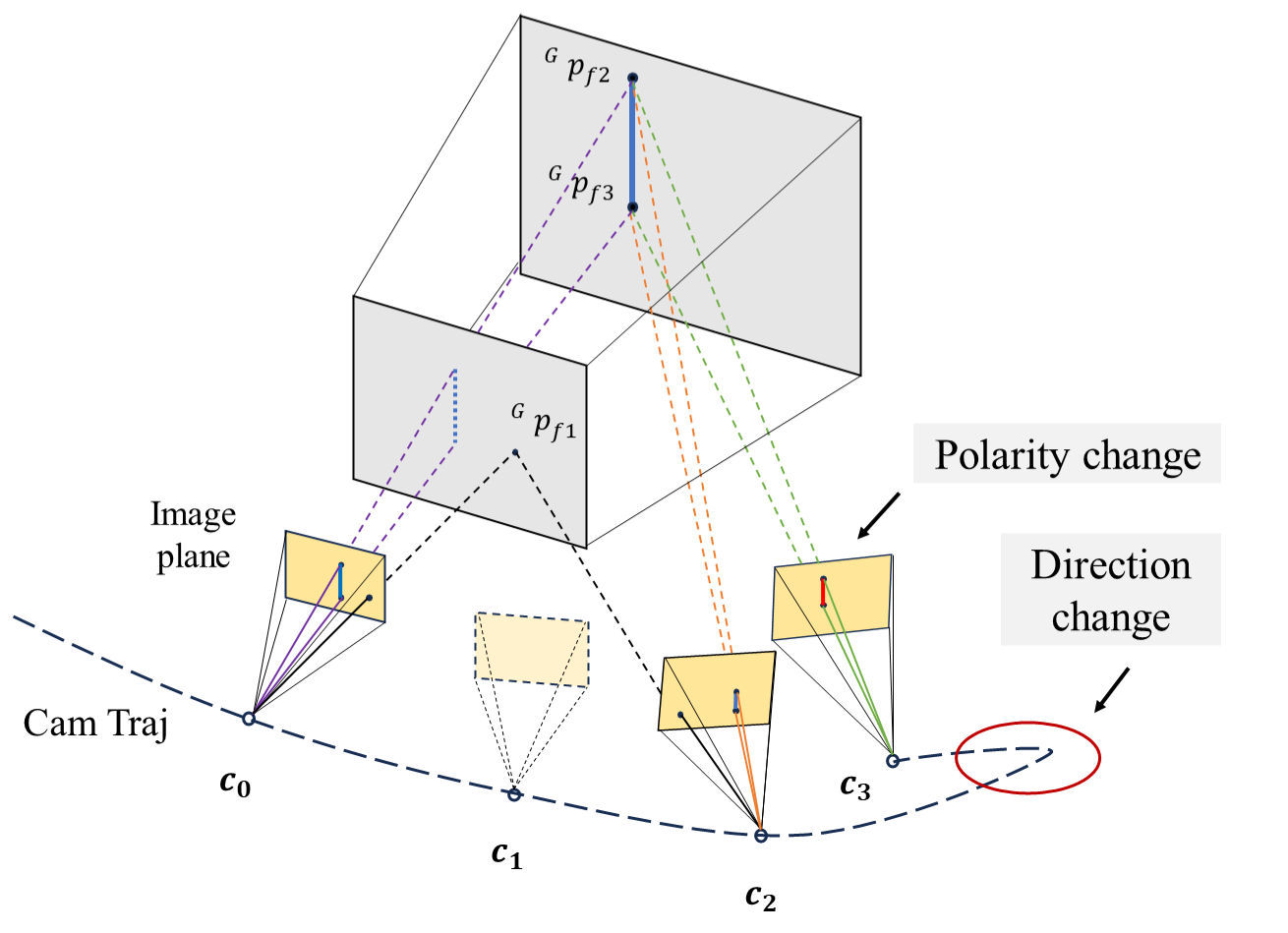}
    \caption{The direction of motion influences the polarity of events. Abrupt changes in motion direction can cause a reversal in event polarity, which may lead to tracking loss. Lines are \textbf{only} used to highlight polarity changes.}
    \label{fig:sec3:tracking-problem}
    \vspace{-1em}
\end{figure}

\smallskip
\noindent
\textbf{Adaptive decay-based Time Surface (ATS) }: We then discuss how to use the adaptive decay $\tau_2(\cdot)$ to generate an adaptive time surface with timestamp $t_j$. The adaptive decay kernel $\tau_2(\cdot)$ includes the event activity which could reflect the dynamic characteristics of the event stream. Thus, the adaptive time surface based on the kernel $\tau_2(\cdot)$ is more robust to various event frequencies and naturally provides a novel and effective way to filter out inactive events and reduce noise. Specifically, any previous event $e_p$ with $\nu(e_p, t_j) < w_{th}$ is considered inactive ($t_j $ is the timestamp of the adaptive time surface), with $\nu(e_p, t)$ is defined as:
\begin{equation}
    \label{eqn:threshold}
    \nu(e_p, t) = \frac{1}{1+r\cdot\alpha(t)(t-t_{p})} \,,
\end{equation}
here $w_{th}$ is a flexible threshold and $\nu(\cdot)$ is modified from Eq.~\eqref{eqn:adaptive_decay_kernel} for the convenience of calculation. In practical computations, $\alpha(t)$ can be approximated by the event activity associated with the closest event to time $t$.

When the system receives an event message, as shown in the Fig.~\ref{fig:sec3:influence_of_activity}, the event activity is calculated recursively, i.e., the decay $\tau_2(e_{k-1}, t_k)$ is calculated first from Eq.~\eqref{eqn:adaptive_decay_kernel}, and then Eq.~\eqref{eqn:event_activity} is used to calculate the event activity $\alpha(t_k)$ at the event timestamp $t_k$. $n(t_k,t_{k-1})$ is always 1 since there is only one event in adjacent moments. In this recursive manner, calculate all event activity at the moment of the event timestamp and save it with the corresponding event.

And there are two different strategies to determine the initial event timestamp $t_{init}$ and the timestamp $t_j$ of the adaptive decay-based time surface: 
\begin{itemize}
    \item \textbf{Data-Priority}~\cite{nunes2023adaptive}: $t_{init}$ starts with the first event timestamp and is typically reset after obtaining a set of active events when $\nu(e_{init}, t_j) < w_{th}$. The method tends to utilize all events without duplication, but the timestamp $t_j$ of the time surface cannot be predicted in advance. In addition, the frequency of the time surfaces generated based on this method is sensitive to the threshold $w_{th}$. 

    \item \textbf{Time-Priority}: we first estimate $t_{init}$ based on the the timestamp $t_j$ of the time surface, event activity $\alpha(t_j)$ and predefined threshold $w_{th}$, as follows:
    \begin{equation}
    \label{eqn:ats_threshold}
    \nu(e_{init},t_j) = w_{th} \implies t_{init} = t_j - \frac{1-w_{th}}{r\cdot\alpha(t_j)w_{th}}\,.
    \end{equation}
    This approach allows for arbitrary timestamps for time surfaces and decouples event activity and generation of the adaptive time surfaces for parallel processing.
\end{itemize}

Events that are considered active have timestamps between $t_{init}$ and $t_j$. We find the latest active events in each pixel, denoted as $e_l(\mathbf{x})$ and the pixel value is $\tau_2(e_l(\mathbf{x}), t_j)$ or $p_l \cdot \tau_2(e_l(\mathbf{x}), t_j)$ if the time surface is polarity-weighted. If the pixel has no active events, set the value of the pixel to zero. Finally, all pixel values are mapped to $[0,255]$.

\subsection{Polarity-aware Feature Tracking}
\label{sec:feature_tracking}

To verify the performance of adaptive decay-based time surfaces in the odometry, we utilize the Fast corners~\cite{rosten2006machine} and track them with the LK optical flow~\cite{lucas1981iterative} method. This approach functions properly within datasets containing event cameras.
However, we observe that abrupt changes in motion direction may lead to tracking failures as they violate the assumption of brightness constancy. Diverse motion directions induce events with opposite polarities at identical scene positions, leading to fluctuations in pixel values, as illustrated in Fig.~\ref{fig:sec1:event_frame} and Fig.~\ref{fig:sec3:tracking-problem}. Such fluctuations notably impact the accuracy of odometry estimation.

Actively inverting event polarities generates a polarity-inverted time surface map, and this map can be regarded as an approximation under the condition of maintaining event polarity, which to some extent mitigates the issue of the violated brightness constancy assumption. By simultaneously tracking both the original and polarity-inverted time surfaces and adaptively merging the tracking results, we ensure continuous and stable feature tracking. Estimating the velocity based on IMU measurements to make this determination may suffer from misjudgments or delays. We track both images separately and then compare the number of successfully tracked features. When the number of successfully tracked features on the polarity-weighted time surface is less than that on the polarity-inverted map, we merge the both tracking results. Finally, we apply the RANSAC method to eliminate outliers and obtain the final matches by estimating the fundamental matrix from the matched features.

\subsection{Update with Tracked Features}
\label{sec:feature_udpate}

When tracking results are available, we select features within the sliding window for triangulation and system update. The system update follows two principles. 
Firstly, we aim to triangulate using as much data as possible to ensure accuracy, thereby ensuring positive gains from system updates. Secondly, stable tracked points are selectively included in the state vector for continuous estimation, aiming to improve system accuracy while maintaining control over increased computation time. 
Based on these principles, features are categorized into SLAM and MSCKF types. 

Specifically, for the current time surface feature tracking, if tracking fails or reaches the maximum length (equal to the sliding window), suggesting that landmark observations are maximized, we attempt to triangulate. 
For a landmark ${}^G\mathbf{p}_f$ associating with a successfully triangulated feature, we have 
\begin{align}
    {}^G\mathbf{p}_f = {}^{I_{j-m}}_G\mathbf{R}^T\left({}^C_I\mathbf{R}^T
    \frac{1}{\rho_l}\pi^{-1}(\begin{bmatrix} u_i \\ v_i \end{bmatrix}) + {}^C\mathbf{p}_I
    \right) + {}^C\mathbf{p}_{I_{j-m}} \,,
\end{align}
where $\pi(\cdot)$ represents the back projection function that transforms a pixel to the normalized image plane. The inverse depth $\rho_l$ is defined in the anchor frame. For instance, the SLAM feature always selects the oldest frame $\{C_{j-m}\}$ in the sliding window as the anchor frame. Consequently, the position ${}^G\mathbf{p}_f$ could be computed using the observation $e_k$ in the anchor frame. 
It’s worth noting that for the MSCKF feature, the anchor frame is determined as the first observed frame.
Given the new observation (e.g. $e_i$) in the latest frame $\{C_{j}\}$, the nonlinear measurement model is given by:
\begin{align}
    \mathbf{z}_{E_i} &= h_e\left(\mathbf{x}_{C_j}, {}^G\mathbf{p}_f\right) + \mathbf{n}_{E_i} = \pi({}^{C_j}\mathbf{p}_f)+\mathbf{n}_{E_i} \\
     {}^{C_j}\mathbf{p}_f &= {}^C_I\mathbf{R}\,{}^{I_{j}}_G\mathbf{R}\left({}^G\mathbf{p}_f - {}^G\mathbf{p}_{I_j}\right) + {}^C\mathbf{p}_I
\end{align}
where the measurement noise is associated with the event pixel noise and follows a white Gaussian distribution $ \mathcal{N}\left(\mathbf{0},\mathbf{Q}_e\right)$.
Features tracked throughout the entire sliding window are augmented into the state vector, continuously updating the system with subsequent landmark observations—referred to as SLAM features. Other successfully triangulated points, termed MSCKF features, are updated using the efficient MSCKF nullspace projection~\cite{mourikis2007multistate}, avoiding the inclusion of landmarks in the state vector and thus reducing system complexity.

\section{Experiments} 
\label{section:experiments}

In this section, we perform extensive experiments to evaluate the efficacy of the proposed method. Initially, we conduct comparisons with classical visual-inertial odometry approaches, specifically two optimization-based methods: ORB-SLAM3~\cite{campos2021orbslam3} and VINS-Mono~\cite{qin2018vins}, along with a filtering-based method, OpenVINS~\cite{genevaopenvins}, on the HKU dataset~\cite{chen2023esvio}. 
Subsequently, we compare our method with event-inertial odometry~\cite{zhu2017eventbased,rebecq2017real,vidal2018ultimate,alzugaray2019asynchronous,guan2022monocular,dai2022tightlycoupled} on the DAVIS 240C dataset~\cite{mueggler2017event}, which is collected using the event camera (with
a resolution of $240 \times 180$) and an internal IMU. Additionally, we conducted ablation experiments on different decay kernels and feature tracking methods, on the dataset~\cite{guan2022monocular}. Both datasets \cite{chen2023esvio, guan2022monocular} contain the DAVIS 346 event camera with a resolution of $346 \times 260$.
Finally, we assess the time consumption of the system to evaluate its real-time performance. The experiments are conducted with the left camera if the stereo event camera is available.

The experiments are carried out on a desktop computer equipped with an Intel Core i7-8700 CPU running Ubuntu 20.04 and ROS Noetic. Accuracy metrics for odometry evaluation consist of Mean Position Error (MPE, \%, per 100 meters), Mean Yaw Error (MYE, deg/m) and Absolute Trajectory Error (ATE, m), while the trajectories are aligned with the ground truth using SE(3) Umeyama alignment~\cite{umeyama1991least} before evaluation. Taking into account the differences between event cameras and to obtain better environmental textures, $r$ in Eq.~\eqref{eqn:adaptive_decay_kernel} is set to 0.2 in the DAVIS 240C dataset and 0.1 for another two datasets.

%%%%%%%%%%%%%%%%%%%%%%%%%%%%%%%%%%%%% visual
\begin{table*}[!ht]
\vspace{5pt}
\captionsetup{font={small}}
% 与视觉惯性里程计的MPE（Unit: %）结果比较。
\caption{Comparison of MPE (Unit: \%) results between the proposed method and visual-inertial odometry. Notably, both VINS-Mono~\cite{qin2018vins} and OpenVINS~\cite{geneva2020openvins} failed across all sequences. Our method demonstrates better adaptability in high-speed and HDR scenes.}
\label{tab:results_with_cam}
\centering
\resizebox{\linewidth}{!}{
\begin{tabular}{c|ccccccccc}
\toprule
Sequence & hku\_agg\_rota & hku\_agg\_small\_flip & hku\_agg\_tran & hku\_agg\_walk & hku\_dark\_normal & hku\_hdr\_agg & hku\_hdr\_circle & hku\_hdr\_slow & hku\_hdr\_tran\_rota \\
\midrule
ORB-SLAM3~\cite{campos2021orbslam3} & {0.711} & {1.895} & {0.841} & {1.389} & {0.695} & {0.623} & {1.451} & {0.635} & {0.971} \\
Ours                                & {0.276} & {0.807} & {0.211} & {0.350} & {0.524} & {0.271} & {0.714} & {0.430} & {0.496} \\
\bottomrule
\end{tabular}
}

% \vspace{-1em}
\end{table*}
%%%%%%%%%%%%%%%%%%%%%%%%%%%%%%%%%%%%% visual

%%%%%%%%%%%%%%%%%%%%%%%%%%%%%%%%%%%%% eio
% \multirow{2}{*}{Sequence}

\begin{table*}[t]
\captionsetup{font={small}}
% 基于DAVIS 240C数据集的里程计平均位置误差（MPE）比较，其中用于参考的方法均为EIO方法。
\caption{Comparison of Odometry MPE (Unit, \%) on the DAVIS 240C dataset~\cite{mueggler2017event}. The methods tested below are all EIO methods. The MPE results are obtained from respective articles, with Alzugaray's results specifically cited from \cite{guan2023evisam}. Because of the differences in the calculation of rotational errors for these methods used for comparison, only MYEs of the proposed method are included. }
\label{tab:results_with_events}
\centering
\resizebox{0.95\linewidth}{!}{

\begin{tabular}{cccccccccc}
\toprule
Sequence  & Length (m) & Zhu's~\cite{zhu2017eventbased} & Rebecq's~\cite{rebecq2017real} & Vidal's~\cite{vidal2018ultimate} & Alzugaray's~\cite{alzugaray2019asynchronous} & Guan's~\cite{guan2022monocular} & Dai's~\cite{dai2022tightlycoupled}  & Ours  \\ \midrule
boxes\_6dof              & {69.852}   & {3.61} & {0.69}              & {\underline{0.44}}   & {2.03}               & {0.61}               & {1.5}     & {\textbf{0.32} (0.02)}   \\
boxes\_translation       & {65.237}   & {2.69} & {0.57}              & {0.76}               & {2.55}               & {\textbf{0.34}}      & {1.0}     & {\underline{0.36} (0.01)}   \\
dynamic\_6dof            & {39.615}   & {4.07} & {0.54}              & {\textbf{0.38}}      & {0.52}               & {\underline{0.43}}   & {1.5}     & {0.49 (0.05)}   \\
dynamic\_translation     & {30.068}   & {1.90} & {\underline{0.47}}  & {0.59}               & {1.32}               & {\textbf{0.26}}      & {0.9}     & {0.59 (0.05)}   \\
hdr\_boxes               & {50.088}   & {1.23} & {0.92}              & {0.67}               & {1.75}               & {\underline{0.40}}   & {1.8}     & {\textbf{0.31} (0.02)}   \\
hdr\_poster              & {55.437}   & {2.63} & {0.59}              & {0.49}               & {0.57}               & {\underline{0.40}}   & {2.8}     & {\textbf{0.18} (0.02)}   \\ 
poster\_6dof             & {61.143}   & {3.56} & {0.82}              & {\underline{0.30}}   & {1.50}               & {\textbf{0.26}}      & {1.2}     & {0.31 (0.03)} \\ 
poster\_translation      & {49.265}   & {0.94} & {0.89}              & {\textbf{0.15}}      & {1.34}               & {0.40}               & {1.9}     & {\underline{0.23} (0.04)} \\ \midrule
Average                  & {52.588}   & 2.58   & 0.69                & 0.47                 & 1.45                 & \underline{0.39}     & 1.58      & \textbf{0.35} (0.03) \\
% shapes\_6dof            & 47.578           & {/} & {/} & {/} & {/} & {0.536 / 0.255}   \\ 
% shapes\_translation     & 56.082           & {/} & {/} & {/} & {/} & {0.383 / 0.215}   \\ 
\bottomrule
\end{tabular}
}
\vspace{-1em}
\end{table*}
%%%%%%%%%%%%%%%%%%%%%%%%%%%%%%%%%% eio

% 里程计精度测试与比较
\subsection{Odometry Accuracy Evaluation and Comparison}

\smallskip
\noindent
\textbf{Comparison with VIO}: 
We first test the accuracy of three visual odometry methods with the proposed odometry to validate the superiority of event cameras in challenging environments. The results are presented in Tab.~\ref{tab:results_with_cam}. VINS-Mono~\cite{qin2018vins} and OpenVINS~\cite{geneva2020openvins} failed in all sequences in this dataset, while ORB-SLAM3~\cite{campos2021orbslam3} (using the monocular visual-inertial method) demonstrates a larger MPE in all sequences, with an average MPE of 1.023. Our method exhibits consistent performance across all sequences with an average MPE of 0.453. From the above results, our event-based odometry is more accurate than conventional camera-based methods in environments with high dynamic ranges and intense motion.

\begin{figure}[ht]
\centering
    \begin{subfigure}[b]{0.42\textwidth}
    \includegraphics[width=\textwidth]{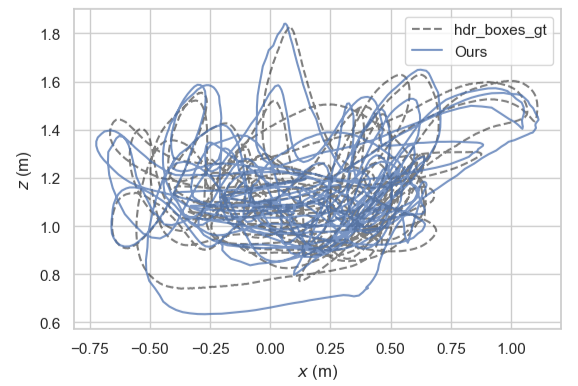}
    \end{subfigure}
    \begin{subfigure}[b]{0.42\textwidth}
    \includegraphics[width=\textwidth]{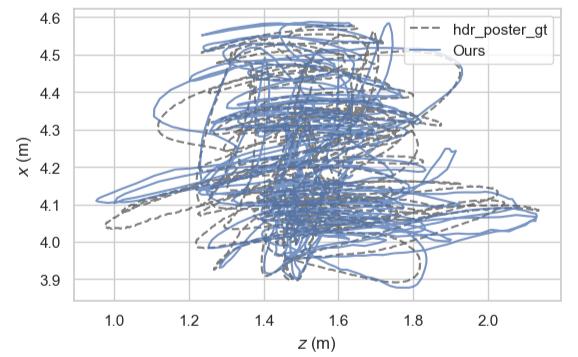}
    \end{subfigure}
    % 误差图，分别属于DAVIS 240C数据集中的$poster_6dof$和$boxes_6dof$序列。
    \caption{Trajectory Estimates Comparison for $hdr\_boxes$ (top) and $hdr\_poster$ (down) in the DAVIS 240C Dataset~\cite{mueggler2017event}}
    \label{fig:sec4:traj_error_graph}
    \vspace{-1em}
\end{figure}

\smallskip
\noindent
\textbf{Comparison with EIO}: 
Next, we compare the accuracy of this method with other event-based approaches. The estimated and ground-truth trajectories were aligned with the subset [5-10]s. The results are presented in Tab.~\ref{tab:results_with_events}. The proposed method achieves an average MPE of 0.35, indicating an error of 0.35m for 100m motion. The test results demonstrate that our method outperforms others in six sequences and shows comparable performance in the remaining sequences. The estimated trajectories of sequence \textit{hdr\_boxes} and \textit{hdr\_poster} are shown in Fig.~\ref{fig:sec4:traj_error_graph}. The figure shows that our odometry can provide good pose estimates, but can still find a partial difference between the estimated trajectory and the ground truth. When the event camera moves slowly, the output events are not sufficient to reflect the texture of the environment, which inevitably leads to inadequate features and inaccurate tracking. Compared to other event-inertial odometry methods, our approach estimates poses more accurately. The adaptive decay-based method enables high-quality texture for feature extraction and tracking, enhancing odometry accuracy with the proposed tracking method.

\subsection{Decay Comparison and Parameter Setting}

In this section, we analyze how different decay functions (exponential and adaptive decay) for time surface representations affect odometry accuracy. 
Specifically, we evaluate various time constants $\eta$ for the exponential decay kernel and compare them with the adaptive decay kernel (as shown in Tab.~\ref{tab:ablation_study}). The experiments utilize a dataset publicly available in~\cite{guan2022monocular}, including a DAVIS 346 event camera.

\smallskip
\noindent
\textbf{Comparison of different decays}: 
 Tab.~\ref{tab:ablation_study} reveals that different time constants $\eta$ affect the accuracy of the odometry. Usually, a larger time constant will keep more information from older events but also introduce unwanted disturbances. Setting $\eta$ to 90ms may result in coarser and overlapping edges, leading to lower precision. However, in sequences such as \textit{vicon\_hdr1} and \textit{vicon\_hdr2}, slightly larger time constants can retain more information, with $\eta$ set to 60ms exhibiting slightly higher accuracy compared to $\eta$ set to 30ms. The odometry with exponential decay-based time surface achieves the relatively best average accuracy when the time constant $\eta$ is 30ms. But it is inevitably prone to bold edges and trailing when the event camera motion suddenly becomes faster. The experimental results show that the odometry with the adaptive decay-based time surface achieves the lowest average ATE of 0.182m. The adaptive decay, which adjusts the decay rate according to the dynamics of the event stream, facilitates the creation of high-quality time surface maps, providing better odometry accuracy than exponential decay.

\begin{figure}[t]
    \centering
    \includegraphics[width=0.95\linewidth]{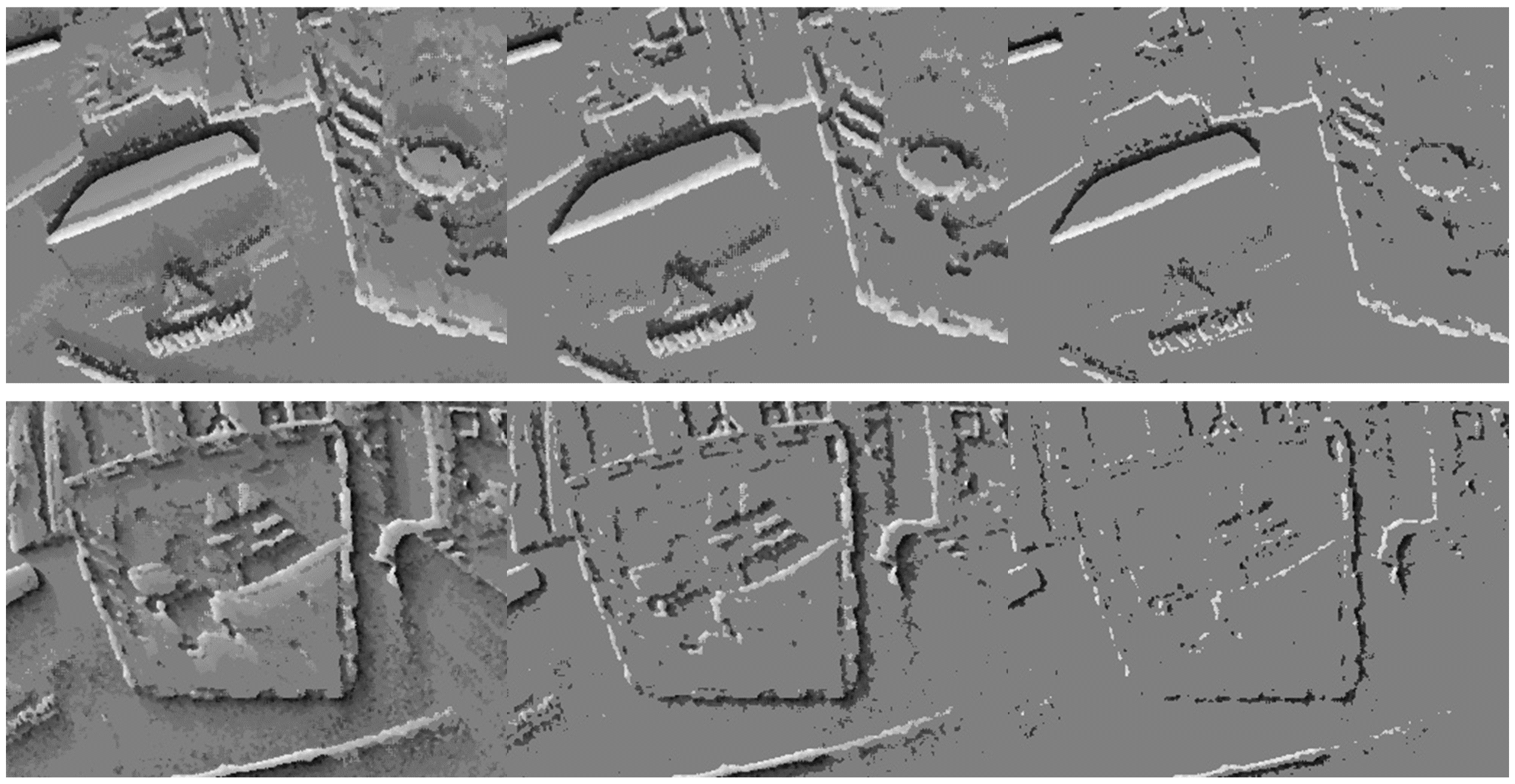}
    %\includegraphics[scale=1.0]{figurefile}
    % 自适应时间面图不同的阈值设置，从左到右的依次是为0.01、0.05和0.1。阈值通常需要合理设置，在保证足够事件数量的同时减少干扰。
    \caption{Comparison of threshold $w_{th}$ settings for adaptive decay-based time surfaces: 0.01, 0.05, and 0.1. A median blur with the kernel size of 1 is used to emphasize the difference. The threshold $w_{th}$ needs to be set reasonably to ensure a sufficient number of events while reducing interference.}
    \label{fig:sec4:ex_atsm}
    \vspace{-1em}
\end{figure}

\smallskip
\noindent
\textbf{Comparison of Different Thresholds $w_{th}$}:
The threshold $w_{th}$ provides a novel solution for filtering out events that contribute little to the texture and saving the time surface generation time. From Eq.~\eqref{eqn:ats_threshold}, $t_{\text{init}}$ is first estimated based on the event activity to bound the time range of active events. If the event activity remains constant, a larger threshold $w_{th}$ tends to utilize events within a smaller time interval, occasionally hindering the formation of desired environmental textures. Conversely, a smaller threshold tends to use events within a larger time interval, enriching image information but introducing interference from old events. Fig.~\ref{fig:sec4:ex_atsm} depicts adaptive time surface maps with different threshold settings. A large threshold of 0.1 results in insufficient texture, while a small threshold of 0.01 incorporates older events that cause trailing effects. 
It requires a reasonable setting of the threshold $w_{th}$ according to practical requirements to achieve optimal texture performance. And a good threshold $w_{th}$ setting usually accommodates all sequences in the dataset without having to set it individually for each sequence. It is important to note that the pixel values for pixels containing active events are determined by the decay kernel and the active event, rather than the threshold $w_{th}$.

%%%%%%%%%%%%%%%%%%%%%%%%%%%%%%% ablation
\begin{table}[t]
\vspace{5pt}
\captionsetup{font={small}}

\caption{Comparison of odometry ATE (m) for different time surface representations. TS stands for exponential decay, while ATS stands for adaptive decay. The parameter $\eta$ denotes the time constant in the exponential decay, tested at values of 30, 60, and 90 ms. The threshold $w_{th}$ for the adaptive decay is consistently set to 0.01, and $(T)$ denotes the improved tracking algorithm.}
\label{tab:ablation_study}
\centering
\resizebox{\linewidth}{!}{
\begin{tabular}{c|ccccc}
\toprule
\multirow{2}{*}{Sequence}   & TS  & TS  & TS    & ATS   & ATS (T)  \\
{}  & {$\eta = 30$} & {$\eta = 60$} & {$\eta = 90$}   & {$w_{th} = 0.01$} & {$w_{th} = 0.01$}  \\ 
\midrule
vicon\_aggressive\_hdr            & {0.377}            & {0.397}              & {0.450}          & {\textbf{0.155}}     & {\underline{0.163}}   \\
vicon\_dark1                      & {0.248}            & {0.294}              & {0.251}          & {\underline{0.124}}  & {\textbf{0.111}}      \\
vicon\_dark2                      & {\textbf{0.120}}   & {\underline{0.163}}  & {0.191}          & {0.253}              & {0.187}               \\
vicon\_darktolight1               & {0.273}            & {0.309}              & {0.359}          & {\textbf{0.177}}     & {\underline{0.219}}   \\
vicon\_darktolight2               & {0.235}            & {0.353}              & {0.249}          & {\underline{0.209}}  & {\textbf{0.187}}      \\
vicon\_hdr1                       & {0.277}            & {\underline{0.263}}  & {0.426}          & {0.267}              & {\underline{0.180}}   \\
vicon\_hdr2                       & {0.365}            & {0.282}              & {0.866}          & {\underline{0.232}}  & {\textbf{0.220}}      \\
vicon\_hdr3                       & {0.187}            & {0.280}              & {0.269}          & {\underline{0.123}}  & {\textbf{0.122}}      \\
vicon\_hdr4                       & {0.229}            & {0.272}              & {0.520}          & {\underline{0.192}}  & {\textbf{0.171}}      \\
vicon\_lighttodark1               & {0.393}            & {0.362}              & {0.405}          & {\underline{0.228}}  & {\textbf{0.226}}      \\
vicon\_lighttodark2               & {0.306}            & {0.501}              & {0.596}          & {\textbf{0.227}}     & {\underline{0.211}}   \\
\midrule
Average & 0.274 & 0.316 & 0.417  & \underline{0.198} & \textbf{0.182} \\
\bottomrule
\end{tabular}
}
\vspace{-1em}
\end{table}
%%%%%%%%%%%%%%%%%%%%%%%%%% ablation 

\subsection{Polarity-aware Feature Tracking Evaluation}

Subsequently, we investigate the effectiveness of the polarity weighting and the polarity-aware feature tracking process, as presented in Tab.~\ref{tab:ablation_study}. Experimental results indicate some improvement when using polarity weighting, achieving average ATE of 0.210m. Polarity weighting enhances the distinction between brighter and darker pixels in the time surface maps, particularly accentuating nearby edges. While offering the above benefits, it also introduces challenges for feature tracking, mainly when the event camera's motion direction changes abruptly, causing shifts in event polarity and undermining the assumption of intensity constancy. Our method incorporates the tracking results of time surfaces with inverted polarity, which somewhat mitigates the problem of tracking failures due to abrupt changes in the direction of motion, and thus achieves an improvement in the accuracy.

\subsection{Computational Efficiency}

Finally, we conduct a detailed analysis of the time consumption for each operation in the proposed method, as outlined in Tab.~\ref{tab:exec_time}. The analysis is conducted on a typical sequence \textit{vicon\_aggressive\_hdr}, which has aggressive motion speed and includes high dynamic range scenes. 
% The time surface generation module operates with eight threads. 
The reported results represent the average values obtained from multiple tests. The statistical results indicate that the average time required for generating an adaptive time surface map is 5.4ms, while the adaptive tracking module is 6.7ms. The total time required for the system to complete one update is 16ms, indicating its capability for real-time performance.

\begin{table}[t]
\vspace{5pt}
\captionsetup{font={small}}
% 各环节的平均执行时间，单位为毫秒。
\caption{Average Execution Time of Each Step in milliseconds}
\label{tab:exec_time}
\centering
\resizebox{\linewidth}{!}{
\begin{tabular}{ccccccc}
\toprule
\begin{tabular}[c]{@{}c@{}}Operation\end{tabular}&
\begin{tabular}[c]{@{}c@{}}ATS\\Generation\end{tabular}&
\begin{tabular}[c]{@{}c@{}}Feature Detection\\\& Tracking\end{tabular}&
\begin{tabular}[c]{@{}c@{}}MSCKF Feature\\Update\end{tabular}&
\begin{tabular}[c]{@{}c@{}}SLAM Feature\\Update\end{tabular}&
\begin{tabular}[c]{@{}c@{}}Others\end{tabular}&
\begin{tabular}[c]{@{}c@{}}Total\end{tabular}
\\ \midrule
% Operation & ATSM\\ Conversion & Point\\ Detection\\ \& Tracking & MSCKF\\ Point\\ Update & SLAM\\ Point\\ Update & Others  & Total \\ \midrule
\begin{tabular}[c]{@{}c@{}}Average \\ Time (ms)\end{tabular}&  
5.4  & 6.7 & 1.3 & 0.2 & 2.4 & 16.0 
\\ \midrule
% Operation     \\ \midrule
% Average Time (ms)       \\ \bottomrule
\end{tabular}
}
\vspace{-1.5em}
\end{table}

\section{Conclusions and Future Work}
\label{sec:conclusion}

Cameras frequently experience motion blur during rapid movement and are constrained by a limited dynamic range. Event cameras, with novel dynamic vision sensors, offer potential solutions to these problems by tracking changes in pixel brightness. However, the asynchronous output of event cameras poses challenges in leveraging event information. 
We propose a monocular event-inertial odometry with an adaptive decay kernel-based time surface and an MSCKF in the back-end for state propagation and update. The adaptive decay highlights recent events according to the dynamic characteristics of the event stream and can also filter out inactive events. Polarity-weighted time surfaces suffer from polarity shifts when motion direction changes abruptly, and we refine feature tracking by incorporating an additional polarity-inverted time surface map to improve robustness. Extensive experiments demonstrate the competitive accuracy of our proposed method. The number of events may decrease during slow motion, resulting in inadequate textures that pose challenges to stable feature tracking. We will continue to work on improving its robustness and accuracy.


%%%%%%%%%%%%%%%%%%%%%%%%%%%%%%%%%%%%%%%%%%%%%%%%%%%%%%%%%%%%%%%%%%%%%%%%%%%%%%%%

%%%%%%%%%%%%%%%%%%%%%%%%%%%%%%%%%%%%%%%%%%%%%%%%%%%%%%%%%%%%%%%%%%%%%%%%%%%%%%%%

%%%%%%%%%%%%%%%%%%%%%%%%%%%%%%%%%%%%%%%%%%%%%%%%%%%%%%%%%%%%%%%%%%%%%%%%%%%%%%%%

%%%%%%%%%%%%%%%%%%%%%%%%%%%%%%%%%%%%%%%%%%%%%%%%%%%%%%%%%%%%%%%%%%%%%%%%%%%%%%%%
%% Original Bib format
% \bibliographystyle{ieeetr}
% \bibliography{icra}

%\clearpage
% {
% New Bib formatll
\AtNextBibliography{\small}
\printbibliography
% }

% \bibliographystyle{IEEEtran}
% \bibliography{refs}

\end{document}